\documentclass[review,3p,times]{elsarticle}
\usepackage{amsmath,amsthm,verbatim,amssymb,amsfonts,amscd,graphicx}

\usepackage[utf8]{inputenc}
\usepackage{longtable}
\usepackage{pgfplots}
\usepackage{multirow}
\usepackage{makecell}
\usepackage{enumitem}
\usepackage{xfrac}
\usepackage[title]{appendix}
\usepackage[english]{babel}
\usepackage{pdflscape}
\usepackage{hyperref}
\usepackage{booktabs}

\usepackage[nodisplayskipstretch]{setspace}

\usepackage[tableposition=top]{caption}  % caption is on top of table
\usepackage{subcaption}
\usepackage{tikz}
\usetikzlibrary{arrows,shapes,positioning,shadows,trees,calc}

\tikzset{
    basic/.style  = {draw, text width=2cm, drop shadow, font=\sffamily, rectangle},
    root/.style   = {basic, rounded corners=2pt, thin, align=center,
                     fill=green!30},
    level 2/.style = {basic, rounded corners=6pt, thin,align=center, fill=green!60,
                     text width=8em},
    level 3/.style = {basic, thin, align=center, fill=pink!60, text width=9.5em}
}

\usepackage[linesnumbered,ruled]{algorithm2e}

\def\R{\mathbb R}

\def\N{\mathbb N}
\def\NN{\mathcal N}

\def\O{\mathcal O}

\def\S{\mathcal S}

\let\emptyset\varnothing

\DeclareMathOperator*{\argmin}{arg\,min}

\newcommand{\eps}{\varepsilon}

\newlength{\summtableheight}  % height of summary tables
\newlength{\tableheight}  % height of each table in appendix
\usepackage{array}
\tikzstyle{linesAll} = [thick, line join=round, mark phase=1, mark repeat=1]

\def\myMarkerLineWidth{0.25}
\def\myMarkSize{1.6}
\def\myMarkSizeD{0.2}
\tikzstyle{markerCirc} = [mark size=\myMarkSize pt, mark=*, mark options={solid, draw=black, line width=\myMarkerLineWidth pt}]
\tikzstyle{markerAst} = [mark size=\myMarkSize pt, mark=asterisk, mark options={solid, draw=black, line width=\myMarkerLineWidth pt}]
\tikzstyle{markerTria} = [mark size=\myMarkSize+2*\myMarkSizeD pt, mark=triangle*, mark options={solid, draw=black, line width=\myMarkerLineWidth pt}]
\tikzstyle{markerSquare} = [mark size=\myMarkSize-\myMarkSizeD pt, mark=square*, mark options={solid, draw=black, line
width=\myMarkerLineWidth pt}]
\tikzstyle{markerDiam} = [mark size=\myMarkSize+\myMarkSizeD pt, mark=diamond*, mark options={solid, draw=black, line
width=\myMarkerLineWidth pt}]
\tikzstyle{markerO} = [mark size=\myMarkSize pt, mark=o, mark options={solid, draw=black, line width=\myMarkerLineWidth pt}]
\tikzstyle{markerMercedes} = [mark size=\myMarkSize+4*\myMarkSizeD pt, mark=Mercedes star, mark options={solid, draw=black, line width=\myMarkerLineWidth pt}]
\tikzstyle{markerStick} = [mark size=\myMarkSize+4*\myMarkSizeD pt, mark=|, mark options={solid, draw=black, line width=\myMarkerLineWidth pt}]
\tikzstyle{plotStyle1} = [linesAll, markerCirc, color=red, solid]
\tikzstyle{plotStyle2} = [linesAll, markerAst, color=green, dashed]
\tikzstyle{plotStyle3} = [linesAll, markerTria, color=blue, dotted]
\tikzstyle{plotStyle4} = [linesAll, markerSquare, color=magenta, dashdotted]
\tikzstyle{plotStyle5} = [linesAll, markerDiam, color=pink, solid]
\tikzstyle{plotStyle6} = [linesAll, markerO, color=teal, dashed]
\tikzstyle{plotStyle7} = [linesAll, markerMercedes, color=violet, dotted]
\tikzstyle{plotStyle8} = [linesAll, markerStick, color=cyan, dashdotted]

\pgfplotsset{compat=1.18}

\journal{Journal of Parallel and Distributed Computing}

\begin{document}

\begin{frontmatter}

%\title{VNS-Accelerated Global  of Sum-of-Squares Clustering for Big Data}
\title{Boosting K-means for Big Data by Fusing Data Streaming with Global Optimization}

\author[a,b]{Ravil Mussabayev\corref{cor1}}
\ead{ravmus@uw.edu}

\cortext[cor1]{Corresponding author.}

\affiliation[a]{organization={Department of Mathematics, University of Washington},
            addressline={Padelford Hall C-138}, 
            city={Seattle},
            postcode={98195-4350}, 
            state={WA},
            country={USA}}

\affiliation[b]{organization={Satbayev University},
            addressline={Satbaev str. 22}, 
            city={Almaty},
            postcode={050013},
            country={Kazakhstan}}

\affiliation[c]{organization={Laboratory for Analysis and Modeling of Information Processes, Institute of Information and Computational Technologies},
            addressline={Pushkin str. 125}, 
            city={Almaty},
            postcode={050010}, 
            country={Kazakhstan}}

\author[b,c]{Rustam Mussabayev}
\ead{rustam@iict.kz}

\begin{abstract}
K-means clustering is a cornerstone of data mining, but its efficiency deteriorates when confronted with massive datasets. To address this limitation, we propose a novel heuristic algorithm that leverages the Variable Neighborhood Search (VNS) metaheuristic to optimize K-means clustering for big data. Our approach is basen on the sequential optimization of the partial objective function landscapes obtained by restricting the Minimum Sum-of-Squares Clustering (MSSC) formulation to random samples from the original big dataset. Within each landscape, systematically expanding neighborhoods of the currently best (incumbent) solution are explored by reinitializing all degenerate and a varying number of additional centroids. Extensive and rigorous experimentation on a large number of real-world datasets reveals that by transforming the traditional local search into a global one, our algorithm significantly enhances the accuracy and efficiency of K-means clustering in big data environments, becoming the new state of the art in the field.
\end{abstract}


\end{frontmatter}

\section{Introduction} \label{sec:big_vns_clust_intro}

The process of clustering represents a foundational function, consisting in detecting clusters of objects that share similarities within a specified object collection. This task is complicated by the swift expansion of digital information and is utilized across a multitude of applied fields (e.g., image analysis, customer segmentation, etc.). Cluster analysis is a multifaceted field with various models, one of which is the Minimum Sum-of-Squares Clustering (MSSC)~\cite{Aloise2009}. At its core, the MSSC is about finding the best way to group a set of $m$ data points within a multi-dimensional space. The goal is to locate $k$ central points, known as centroids, in such a way that the total squared distances from each data point to the nearest centroid are minimized.

Mathematically, this is expressed as:

\begin{equation}
	\min\limits_{C} \ \ \ f\left(C,X\right)=\sum\limits_{i=1}^m \min_{j=1,\ldots,k} \| x_i - c_j \|^2
	\label{eq:mssc}
\end{equation}

Here, the challenge is to find the optimal cluster centers $C$ that minimize the distance equation, where $\| \cdot \|$ denotes the Euclidean norm. The equation above defines the main task of the MSSC, called the sum-of-squared distances. When solved, each solution creates a specific partitioning of the data points into clusters.

It is worth noting that the MSSC problem is far from simple. In fact, MSSC is classified as an NP-hard problem~\cite{Aloise2009}. This complexity makes it a global optimization problem with the overarching objective of dividing a dataset into distinct groups or clusters.

One of the standout features of MSSC is how it operates. By minimizing the expression in Equation~\eqref{eq:mssc}, it automatically reduces the similarity within the same cluster and increases the difference between separate clusters. This dual-action approach makes MSSC highly effective and provides an essential measure of clustering accuracy. It not only helps in forming cohesive groups but also ensures that these groups are as distinct from each other as possible, which is often the fundamental objective in clustering algorithms.

Studies have demonstrated that global minimizers offer a more precise representation of the underlying clustering patterns within a dataset~\cite{Gribel2019}. However, finding these global minimizers within the context of MSSC is no easy task due to the complex, non-convex and non-smooth nature of the objective function.

To tackle this difficulty, various strategies have been devised. These include gradient-based optimization, stochastic optimization algorithms, and metaheuristic search strategies, among others. Each of these approaches has its particular focus:
\begin{itemize}
	\item Gradient-based techniques tend to reach local minimizers quickly but can become stuck in suboptimal solutions, owing to the non-convex form of the objective function;
	
	\item Stochastic optimization algorithms, on the other hand, use randomness to break free from local minima, thus exploring a more extensive part of the solution space;
	
	\item Metaheuristic search strategies attempt to find a middle ground between exploring new possibilities and exploiting known solutions;
	
	\item Hybrid methods combining the above techniques in various ways are often employed to capitalize on the strengths of each. This enables them to uncover unique benefits over individual approaches.
\end{itemize}

Despite the array of optimization methods available, no single approach has emerged as the definitive solution for this highly non-convex challenge. Each has its unique strengths and shortcomings. This state of affairs emphasizes the ongoing need for further research in this area. Developing more efficient and effective methods for finding global minimizers is not just a theoretical concern; it is a practical necessity in the expanding field of pattern recognition for big data. By recognizing the complexity of this issue and acknowledging that there is no universal solution, we open the door for innovation and advancement in clustering techniques.

In this work, we present an algorithm, BigVNSClust, whose core concept is to conduct a simultaneous search across two modalities: (1) exploring partial solution landscapes derived from random samples of the original dataset, and (2) cycling through increasingly expansive neighborhoods within these landscapes to refine the incumbent solution. For the latter, a special neighborhood structure can be defined, which defines two solutions to be neighboring if they differ only in a certain number of centroids. Exploring this structure according to some suitable metaheuristic may give enough guidance to achieve a more educated traversal of solutions.

Exploiting both of these modalities allows the algorithm to effectively shake the incumbent solution, and thereby escape unfavorable local minima. By gaining control over the amount of input data we feed into the problem, the algorithm efficiently reduces the time complexity and adaptively scales to big data. Moreover, integrating an advanced metaheuristic technique, Variable Neighborhood Search (VNS)~\cite{Mladenovic1997}, into the search within each consecutive solution landscape enables a more detailed exploration of its local characteristics and peculiarities, including such pitfalls as degenerate clusters or fragmentation of a single true cluster by several centroids.

More specifically, one of the critical drawbacks of the existing initialization-sensitive iterative approaches, like K-means++, lies in their inability to handle the situation when two or more initial centroids fall inside a single cluster that is well-separated from the other data points by a large margin. This occurs due to difficulties for centroids in traveling across the empty space of the margin by shifting centroids to the means, even if the original input dataset is sparsified well enough by drawing a small random sample. We empirically demonstrate this issue in Section (...) of the article.

Thus, in each iteration of the algorithm, the degenerate clusters and $p$ random centroids of the incumbent solution $C$ are reinitialized using K-means++ on a new random sample $S$ from $X$. This initial step amounts to shaking the incumbent solution. Variable $p$ is called the shaking power. It is cyclically incremented across iterations, being bounded by the hyperparameter $p_{max}$. Then, the local search (K-means) is commenced on $S$ from the incumbent solution, seeking a possibly better one in the current neighborhood. The iterations proceed until a predefined time limit $T$.

All in all, our study investigates the potential benefits of combining big data clustering with a cutting-edge optimization metaheuristic, exploring the possibilities of hybridization to enhance clustering outcomes. We introduce a novel clustering heuristic named BigVNSClust, which seeks to amplify the global optimization capabilities of standard approaches. By integrating a powerful metaheuristic framework, Variable Neighborhood Search (VNS)~\cite{Mladenovic1997}, we develop a refined approach to clustering. Through a comprehensive experimental analysis performed on a large number of real-world datasets, BigVNSClust demonstrates state-of-the-art performance within the family of MSSC clustering algorithms. Notably, it surpasses its closest competitor, Big-means~\cite{Mussabayev2023}, solidifying its position as a robust and innovative advancement in the field.

This paper has the following outline. Section~\ref{sec:big_vns_clust_vns} provides an elementary introduction to the VNS metaheuristic. Section~\ref{sec:big_vns_clust_proposed_algo} gives a precise pseudocode of BigVNSClust, discusses its main properties, as well as analyzes the resulting time complexity. Section~\ref{sec:big_vns_clust_experiments} describes the conducted experiments and analyzes the obtained results. Finally, Section~\ref{sec:big_vns_clust_conclusion} concludes the chapter with final thoughts.

\section{Variable Neighborhood Search} \label{sec:big_vns_clust_vns}

\subsection{Main concepts}

In the realm of computer science, artificial intelligence, and mathematical optimization, heuristics represent essential tools designed to expedite problem solving. They are used when conventional methods prove too slow or when the quest for an exact solution becomes intractable. It is crucial to acknowledge, however, that heuristics do not provide an absolute guarantee of uncovering the optimal solution, often falling under the category of approximate algorithms. Typically, these algorithms excel at swiftly and efficiently generating solutions that closely approximate the optimal one. On occasion, they may even attain highest accuracy in identifying the absolute best solution. Nevertheless, they retain their heuristic classification until the solutions produced by them are proved to be optimal~\cite{Desale2015}. In a broader context, metaheuristics serve as versatile frameworks for crafting heuristics to tackle a diverse array of combinatorial and global optimization challenges~\cite{Hansen2014}.

An optimization problem can be generally formulated as
\begin{equation}
	\min \{ f(x) \mid x \in S \subseteq \S \},
\end{equation}
where:
\begin{itemize}
	\item $\S$ represents the ambient solution space;
	\item $S$ denotes the feasible set within $\S$;
	\item $x$ is a feasible solution in $S$;
	\item $f$ is a real-valued objective function.
\end{itemize}
Henceforth, the symbol $S$ is used to represent two distinct concepts: firstly, a feasible solution set $S$ within the ambient solution space $\S$; and secondly, a specific sample $S$ extracted from the entire dataset $X$. The particular interpretation of $S$ in any given instance should be readily discernible to the reader based on the surrounding context since the feasible solution space is $S = \S = \R^n$ in the context of MSSC~\eqref{eq:mssc}.

Variable Neighborhood Search (VNS), introduced by Mladenovic in 1997~\cite{Mladenovic1997}, represents a contemporary metaheuristic approach offering a versatile framework for effectively addressing combinatorial and continuous non-linear global optimization problems. VNS systematically exploits the idea of neighborhood change, both in descent to local minima and in escape from the valleys containing them~\cite{Mladenovic1997, Hansen2003, Hansen2014}. It explores distant neighborhoods of the current solution and moves to a new one if and only if this movement leads to an improvement in the objective function.

VNS (Variable Neighborhood Search) stands as a potent instrument for the construction of novel global search heuristics, leveraging already existing local search heuristics. Its efficacy is underpinned by the following key insights:

\begin{enumerate}[align=left] \label{enum:vns_facts}
	\item [\textbf{Fact 1:}] A solution that constitutes a local optimum within one neighborhood structure does not necessarily remain optimal when analyzed under a different neighborhood;
	\item [\textbf{Fact 2:}] A solution that achieves global optimality will simultaneously be a local optimum across all conceivable neighborhood structures;
	\item [\textbf{Fact 3:}] For a broad spectrum of problems, local optima, when observed in the context of one or multiple neighborhoods, often lie in close proximity to one another.
\end{enumerate}

Fact 1 paves the way for the adoption of intricate moves that aim to discern local optima across all the neighborhood structures in play. Fact 2 implies that expanding the number of neighborhoods might be a strategic move, especially if the current local optima identified are not of satisfactory quality. When combined, these insights highlight an intriguing prospect: a solution recognized as a local optimum across multiple neighborhood structures has a heightened likelihood of achieving global optimality compared to one that is optimal within a single neighborhood structure~\cite{Hansen2016}.

Fact 3, primarily rooted in empirical findings, indicates that a local optimum can often shed light on the nature of the global optimum. Put differently, local optima in optimization problems tend to exhibit similarities among themselves. This consistency underscores the merit of intensively exploring the vicinity of a given solution. To elucidate further, if we collect all local solutions $C = (c_1, \ldots, c_k)$ of the MSSC problem as described in equation~\eqref{eq:mssc} into a single set, then it is plausible that a substantial subset of these locally optimal centroids will be very close to each other. In the meantime, a few might deviate. This pattern hints that the global optimum will bear resemblances in certain variables (specific coordinates of the vector representations) with those found in local optima. Identifying these overlapping variables beforehand is typically elusive. As a result, a systematic exploration of the neighborhoods surrounding the current local optimum is prudent until a superior solution emerges.

VNS comprises two phases: the improvement phase, which refines the current solution by potentially descending to the nearest local optimum, and the shaking (or perturbation) phase, designed to mitigate local minima traps. VNS operates by alternately executing the improvement and shaking procedures, along with a neighborhood change step, until certain predefined stopping criteria are met. Specifically, VNS revolves around three core, alternating steps:
\begin{enumerate}
	\item Shaking procedure;
	\item Improvement procedure;
	\item Neighborhood change step.
\end{enumerate}

Now, let us define some essential notions for the VNS framework.

The incumbent solution is referred to as the best obtained solution $x$ with respect to the objective function value.

A neighborhood of a given solution $x$ refers to the set of solutions that can be directly obtained from $x$ by applying a specific local change. Usually, a neighborhood is constructed based on a metric (or quasi-metric) function denoted as $\delta$. Given a non-negative distance value $\Delta \ge 0$, the neighborhood of a point $x$ can be defined as:
\begin{equation}
	\NN_\Delta(x) = \left\{ y \in S \mid \delta(x, y) \le \Delta \right\}
\end{equation}
For instance, in the context of a continuous optimization problem over $\mathbb{R}^n$, $\NN_1(x)$ may represent an Euclidean ball of radius $1$ centered at $x$. Alternatively, $\NN_3(x)$ could denote the set of all vectors derived from $x$ by swapping (or changing) exactly three coordinates of $x$. In the Traveling Salesman Problem (TSP), one type of neighborhood might consist of all the tours that can be obtained by reversing a subsequence of a given tour. This specific operation is called a ``2-opt'' move.

A neighborhood structure $\NN$ refers to an ordered collection of operators
$$
\NN = \{ \NN_1, \ldots, \NN_{k_{\max}} \},
$$
where each operator $\NN_k: S \to 2^S$ ($2^S$ denotes the power set of $S$), for $k \in \{1, \ldots, k_{\max}\}$, maps a solution $x$ to the predefined neighborhood $\NN_k(x)$. The hierarchy of these neighborhoods plays a pivotal role in VNS. When a local search within one neighborhood fails to find any improvement, VNS typically progresses to the subsequent neighborhood in the sequence, aiming to discover superior solutions. It is common for VNS to employ multiple neighborhood structures throughout its search process, adjusting the neighborhood dynamically during the search to sidestep local optima. In the ensuing discussions, both the collection of operators $\NN$ and the collection of neighborhoods $\NN(x)$ will be referred to as a neighborhood structure.

Each neighborhood structure typically has a different way of defining neighbors. For instance, in the context of the TSP as previously discussed, one could utilize a neighborhood structure based on "$k$-opt" moves. In a "$k$-opt" move, $k$ edges are removed from the tour and subsequently reconnected in a manner different from the original configuration.

We adopt two distinct notations, $\NN$ for the neighborhood structures in the shaking procedure and $N$ for those in the improvement procedure, to differentiate between them.

\subsection{Shaking procedure} \label{subsec:big_vns_clust_shaking}

A shaking procedure is a necessary step that allows to escape local optima traps.

The most straightforward shaking procedure involves selecting a random solution from the neighborhood $\NN_k(x)$, with $k$ being predetermined. For certain problems, a completely random jump within the $k$-th neighborhood may result in an overly aggressive perturbation. As a consequence, an "intensified shaking" approach is sometimes favored. This method considers the sensitivity of the objective function to minor variations in the variable $x$. Nevertheless, for the purposes of this chapter, our definition of the shaking procedure will align with the simpler, random-selection approach. The corresponding pseudocode for this shaking procedure is provided in Algorithm~\ref{alg:shaking_proc}.

\begin{algorithm}
	\SetKwInOut{Input}{Input}
	\SetKwInOut{Output}{Output}
	
	\SetKwFunction{FMain}{Shake}
	\SetKwProg{Fn}{Function}{:}{}
	\Fn{\FMain{$x$, $k$, $\NN$, $P$}}{
		\Input{Incumbent solution $x$; \\
			Neighborhood index $k$; \\
			Neighborhood structure $\NN$; \\
			Probability distribution $P$;
		}
		\Output{Shaken solution $x'$.}
		Choose $x' \in \NN_k(x)$ at random according to $P$; \\
		\Return $x'$
	}
	\caption{Shaking Procedure}
	\label{alg:shaking_proc}
\end{algorithm}

\subsection{Neighborhood change step} \label{subsec:big_vns_clust_nbhd_change}

The neighborhood change procedure aims to steer the direction of the variable neighborhood search heuristic throughout the solution space. Specifically, it dictates which neighborhood will be probed next and determines whether a given solution should be adopted as the new incumbent. While various versions of the neighborhood change step can be found in the literature~\cite{Hansen2016}, the sequential and cyclic methods are the most prevalent.

\begin{enumerate}
	\item Sequential neighborhood change. The steps of this procedure are listed in Algorithm~\ref{alg:seq_nbhd_change}. If an improvement of the incumbent solution in some neighborhood occurs, then the search is resumed in the first neighborhood (according to the predefined order in $\NN$) of the neighborhood structure at the new incumbent solution; otherwise, the search is continued in the subsequent neighborhood of the incumbent solution.
	
	\begin{algorithm}
		\SetKwInOut{Input}{Input}
		\SetKwInOut{Output}{Output}
		
		\SetKwFunction{FMain}{Neighborhood\_change\_sequential}
		\SetKwProg{Fn}{Function}{:}{}
		\Fn{\FMain{$x$, $x'$, $k$}}{
			\Input{Incumbent solution $x$; \\
				Candidate solution $x'$; \\
				Neighborhood index $k$.
			}
			\eIf{$f(x') < f(x)$}{
				$x \gets x'$\tcp*[l]{move}
				$k \gets 1$\tcp*[l]{return to first neighborhood}
			}
			{
				$k \gets k + 1$\tcp*[l]{proceed to next neighborhood}
			}
		}
		\caption{Sequential Neighborhood Change Step}
		\label{alg:seq_nbhd_change}
	\end{algorithm}
	
	\item Cyclic neighborhood change. In this step, regardless of whether there has been an improvement with respect to some neighborhood or not, the search is continued in the subsequent neighborhood (see Algorithm~\ref{alg:cyc_nbhd_change}):
	
	\begin{algorithm}
		\SetKwInOut{Input}{Input}
		\SetKwInOut{Output}{Output}
		
		\SetKwFunction{FMain}{Neighborhood\_change\_cyclic}
		\SetKwProg{Fn}{Function}{:}{}
		\Fn{\FMain{$x$, $x'$, $k$}}{
			\Input{Incumbent solution $x$; \\
				Candidate solution $x'$; \\
				Neighborhood index $k$.
			}
			\If{$f(x') < f(x)$}{
				$x \gets x'$\tcp*[l]{move}
			}
			$k \gets k + 1$\tcp*[l]{proceed to next neighborhood}
		}
		\caption{Cyclic Neigborhood Change Step}
		\label{alg:cyc_nbhd_change}
	\end{algorithm}
	
\end{enumerate}

\subsection{Improvement procedures} \label{subsec:big_vns_clust_improv_proc}

There are two main types of improvement procedures that can be used inside a VNS heuristic: local search and variable neighborhood descent (VND)~\cite{Hansen2016}.

A local search heuristic offers a fundamental approach to solution improvement. At each iteration, it delves into the neighborhood structure $N(x)$ of the incumbent solution $x$. The process starts with this incumbent solution. If a superior solution within the predefined neighborhood of $N(x)$ is identified, it then replaces the incumbent. This iterative refinement continues until reaching a solution that is locally optimal with respect to its neighborhood, signifying no further enhancements are possible.

Two prevalent strategies to navigate the neighborhood $N(x)$ are:
\begin{enumerate}
	\item First Improvement: This strategy immediately adopts the first detected improved solution within $N(x)$ as the new incumbent;
	\item Best Improvement: Here, the most optimal solution amongst all potential improvements in $N(x)$ becomes the new incumbent.
\end{enumerate}
For the scope of this work, our primary focus is on the local search using the best improvement strategy. The corresponding pseudocode is detailed in Algorithm~\ref{alg:bi_loc_search}.

\begin{algorithm}
	\SetKwInOut{Input}{Input}
	\SetKwInOut{Output}{Output}
	
	\SetKwFunction{FMain}{Best\_improvement\_local\_search}
	\SetKwProg{Fn}{Function}{:}{}
	\Fn{\FMain{$x$, $N$}}{
		\Input{Incumbent solution $x$; \\
			Neighborhood $N$;
		}
		\Output{New incumbent solution $x'$.}
		\Repeat{$f(x') \le f(x)$}{
			$x' \gets x$\tcp*[l]{remember old solution}
			$x \gets \argmin_{y \in N(x')} f(y)$\tcp*[l]{find best improving solution}
		}
		\Return $x'$
	}
	\caption{Local Search using Best Improvement}
	\label{alg:bi_loc_search}
\end{algorithm}

A Variable Neighborhood Descent (VND) procedure delves into multiple neighborhoods of the incumbent solution in every iteration. By operating in either a sequential or nested manner, its objective is to refine the current solution. In essence, VND capitalizes on the initial two VNS characteristics previously mentioned. While the choice between the ``first improvement'' and the ``best improvement'' search strategies remains flexible in VND, the basic sequential VND (B-VND) procedure expands the concept of local search. In this context, if a neighborhood structure encompasses several neighborhoods denoted by \( N = \{ N_1, \ldots, N_{l_{\max}} \} \) (with \( l_{\max} \) setting the upper limit on the number of neighborhoods), B-VND systematically investigates these neighborhoods based on this predetermined sequence. When an enhancement in the incumbent solution emerges within a particular neighborhood, the search reinitiates from the foremost neighborhood but pivots around this newly improved solution. The entire procedure culminates when no further refinements can be achieved for the incumbent solution across all \( l_{\max} \) neighborhoods. The procedural steps of the sequential VND, adopting the best improvement search strategy, are detailed in Algorithm~\ref{alg:best_imp_seq_vnd}.

\begin{algorithm}
	\SetKwInOut{Input}{Input}
	\SetKwInOut{Output}{Output}
	
	\SetKwFunction{FMain}{B\_VND}
	\SetKwProg{Fn}{Function}{:}{}
	\Fn{\FMain{$x$, $l_{\max}$, $N$}}{
		\Input{Incumbent solution $x$; \\
			Maximum neighborhood index $l_{\max}$; \\
			Neighborhood structure $N$;
		}
		\Output{New incumbent solution $x'$.}
		\Repeat{$stop = \text{True}$}{
			$stop \gets \text{False}$\;
			$l \gets 1$\;
			$x' \gets x$\;
			\Repeat{$l = l_{\max}$}{
				$x'' \gets \argmin_{y \in N_l(x)} f(y)$\tcp*[l]{find best improving solution}
				$\text{Neighborhood\_change}(x, x'', l)$\tcp*[l]{any neighborhood change function}
			}
			\If{$f(x') \le f(x)$}{
				$stop \gets \text{True}$\;
			}
		}
		\Return $x'$
	}
	\caption{Sequential VND using Best Improvement Search}
	\label{alg:best_imp_seq_vnd}
\end{algorithm}

Also, there is a plethora of other VND procedures: pipe VND (P-VND), cyclic VND (C-VND), union VND (U-VND), nested VND, mixed VND, etc.~\cite{Hansen2016} However, since they were not used in the algorithm proposed in this article, we omit their description and refer the reader to~\cite{Hansen2016} for more details.

\subsection{Basic Variable Neighborhood Search}

Unlike many other metaheuristics, Variable Neighborhood Search (VNS) and its various extensions are characterized by their simplicity, often requiring minimal or even zero tuning of parameters. Consequently, besides delivering high-quality solutions, often in more straightforward fashion compared to alternative approaches, VNS also offers insights into the underlying reasons for such effectiveness. These insights can, in turn, pave the way for the development of more efficient and refined implementations~\cite{Hansen2014}.

\begin{figure}[htbp]
	\centering
	\includegraphics[width=\textwidth]{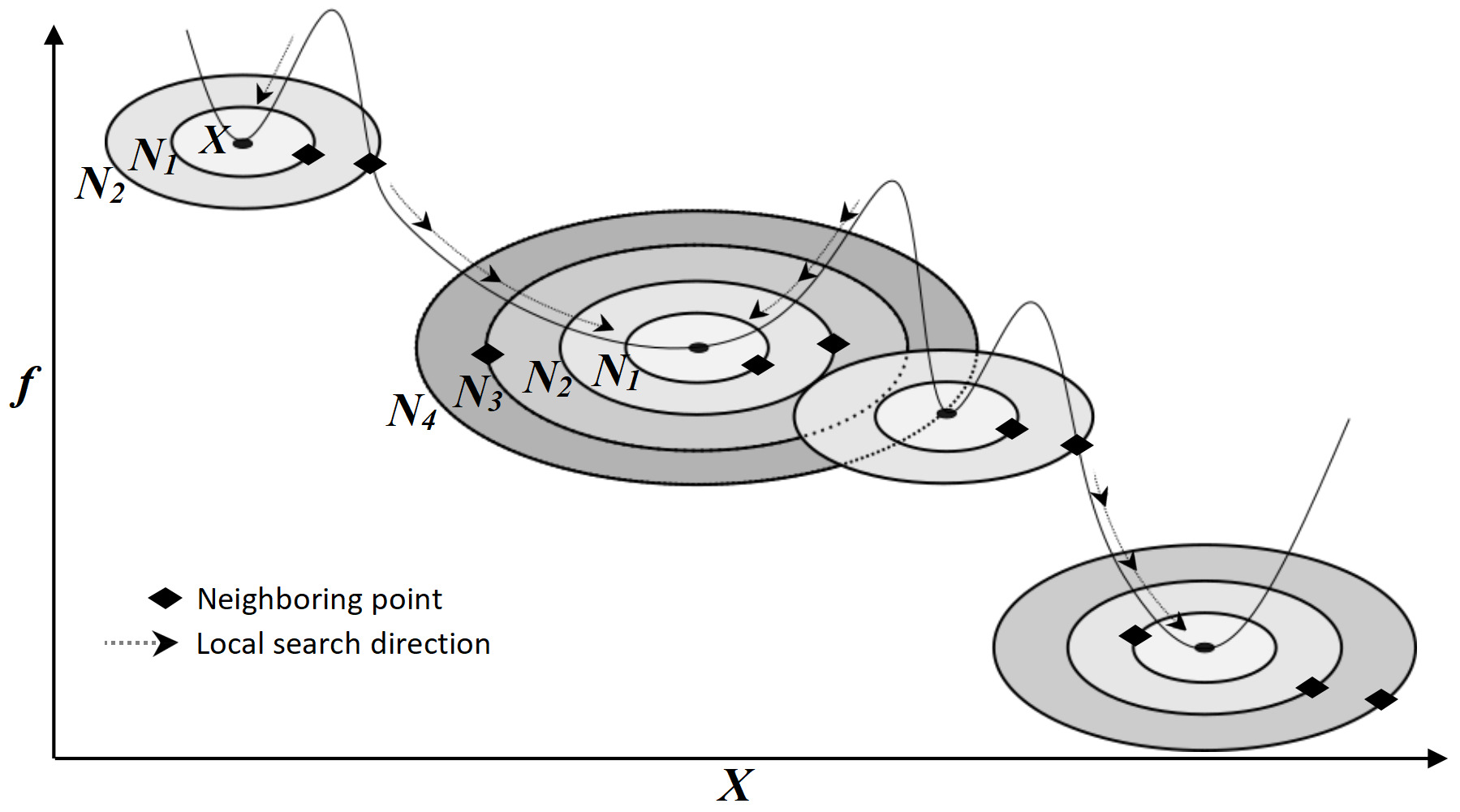}
	\caption{Illustration of VNS applied to minimize some function $f$. Reproduced from ``Constructing the Neighborhood Structure of VNS Based on Binomial Distribution for Solving QUBO Problems'' (Pambudi and Kawamura, 2022) \cite{Pambudi2022}, Algorithms, CC BY 4.0. \href{https://creativecommons.org/licenses/by/4.0/}{https://creativecommons.org/licenses/by/4.0/}}
	\label{fig:vns}
\end{figure}

The pseudocode of the basic VNS is presented in Algorithm~\ref{alg:basic_vns}, while Figure~\ref{fig:vns} schematically shows the process of VNS optimization applied to some non-convex function $f$.

\begin{algorithm}
	\SetKwInOut{Input}{Input}
	\SetKwInOut{Output}{Output}
	
	\SetKwFunction{FMain}{Basic\_VNS}
	\SetKwProg{Fn}{Function}{:}{}
	\Fn{\FMain{$x$, $k_{\max}$, $T$, $\NN$, $N$}}{
		\Input{Initial solution $x$; \\
			Maximum shaking power $k_{\max}$; \\
			Maximum time $T$; \\
			Neighborhood structure for shaking $\NN$; \\
			Neighborhood structure for improvement $N$;
		}
		\Output{Final solution $x$.}
		\Repeat{$\text{CpuTime}() > T$}{
			$k \gets 1$ \tcp*[l]{initially, use the minimal shaking power}
			\While{$k \le k_{\max}$}{
				$x' \gets \text{Shake}(x, k, \NN)$\tcp*[l]{generate $k$-th power variation of incumbent solution}
				$x'' \gets \text{Local\_search}(x', N)$\tcp*[l]{apply some local search method with $x'$ as initial solution}
				$\text{Neighborhood\_change}(x, x'', k)$\tcp*[l]{apply one of described neighborhood change methods}
			}
		}
		\Return $x$
	}
	\caption{Basic Variable Neighborhood Search}
	\label{alg:basic_vns}
\end{algorithm}

Within the basic VNS framework, the following key components can be identified:

\begin{itemize}
	\item $f(x)$: Real-valued objective function;
	\item $k$: Shaking intensity;
	\item $k_{\max}$: Maximum shaking intensity;
	\item $\NN_k(x)$: The $k$-th neighborhood of solution $x$;
	\item $x$: Incumbent (current) solution;
	\item $x' \gets \text{Shake}(x, k, \NN)$: A procedure, detailed in Section~\ref{subsec:big_vns_clust_shaking}, that generates a candidate solution $x'$ randomly selected from the $k$-th neighborhood $\NN_k(x)$ of the incumbent solution $x$;
	\item $\text{NeighborhoodChange}(x, x'', k)$: A procedure for progressively altering the neighborhood, as described in Section~\ref{subsec:big_vns_clust_nbhd_change};
	\item $x'' \gets \text{LocalSearch}(x', N)$: A local search method, elaborated upon in Section~\ref{subsec:big_vns_clust_improv_proc}, utilized to find solutions within the vicinity of local optima;
	\item $\text{CpuTime}()$: Function that returns the time elapsed since the algorithm's initiation;
	\item $T$: A predefined, relatively short time limit for the search.
\end{itemize}

\section{Proposed algorithm} \label{sec:big_vns_clust_proposed_algo}

\subsection{Precise description}

The pseudocode of the proposed BigVNSClust algorithm is shown in Algorithm~\ref{alg:big_vns_clust}. In the context of MSSC, where $k$ already denotes the number of clusters, we use $p$ and $p_{\max}$ to represent the current and maximum shaking powers, respectively.

\begin{algorithm}
	\SetKwInOut{Input}{Input}
	\SetKwInOut{Output}{Output}
	
	\SetKwFunction{FMain}{BigVNSClust}
	\SetKwProg{Fn}{Function}{:}{}
	\Fn{\FMain{$X$, $k$, $s$, $p_{\max} = 3$}}{
		\Input{
			Feature set $X = \{x_1, \ldots, x_m\}$; \\
			Number of clusters $k$; \\
			Sample size $s$; \\
			Maximum shaking power $p_{\max}$ (default = 3);
		}
		\Output{
			Cluster centers $C = \{c_1, \ldots, c_k\}$; \\
			Cluster assignments $Y = \{y_1, \ldots, y_m\}$; \\
			Objective function value $f(C, X)$.
		}
		Initialize centroids $C$ as degenerate clusters $\left\{\emptyset_1, \ldots, \emptyset_k\right\}$\;
		Set initial objective value $\hat{f} \gets \infty$\;
		Initialize shaking power $p \gets 1$ and iteration counter $t \gets 0$\;
		
		\While{$t < T$}{
			Draw a random sample $S$ of size $s$ from $X$\;
			Randomly select $p$ centroids from $C$ as $C'$\;
			
			\For{each centroid $c \in C$}{
				\If{$c \in C'$ or $c$ is empty}{
					Reinitialize $c$ using K-means++ on $S$\;
				}
			}
			
			Update centroids $C_{\text{new}}$ by applying K-means on $S$ starting with $C$\;
			
			\If{$f(C_{\text{new}}, S) < \hat{f}$}{
				$C \gets C_{\text{new}}$\;
				$\hat{f} \gets f(C_{\text{new}}, S)$\;
			}
			
			Increase shaking power: $p \gets p + 1$\;
			\If{$p > p_{\max}$}{$p \gets 1$}
			
			$t \gets t + 1$\;
		}
		
		Assign each point in $X$ to the nearest centroid in $C$ to compute $Y$\;
		\Return $C, \ \ Y, \ \ f(C, X)$
	}
	\caption{Variable Neighborhood Search (VNS) for Big Data Clustering}
	\label{alg:big_vns_clust}
\end{algorithm}

The proposed Algorithm~\ref{alg:big_vns_clust} follows the basic VNS variation, as described in Algorithm~\ref{alg:basic_vns}, using the following choice of ingredients:
\begin{enumerate}
	\item Neighborhood structure $\NN$ employed by the shaking procedure is defined as follows. Let $S \subset X$ be a uniform random sample of size $s$ from $X$. Also, let $U = (u_1, \ldots, u_l) \in \R^{l \times n}$ be a finite set of points in $X$. Define a probability distribution $P: X \to [0, 1]$ by
	\begin{equation}
		P_U(x) = \frac{ d(x, U) }{ \sum_{x' \in X} d(x', U) }  \label{sampling_distr}
	\end{equation}
	where $d(x, U)$ denotes the distance from point $x$ to set $U$. The distribution $P$ is exactly the distribution used in the K-means++ seeding. Let $C = (c_1, \ldots, c_k)$ be the incumbent solution. Then, the $p$-th neighborhood $\NN_p(C)$ of incumbent solution $C$ is defined as follows:
	\begin{align}
		\NN_p(C) = \{ (& c_1, c_2, \ldots, c_{l_1}, \ldots, c_{l_2}, \ldots, c_{l_p}, \ldots, c_k) \ | \\
		& \text{where} \ l_1, l_2, \ldots, l_p \in \N_k \ \text{such that} \ l_i \ne l_j \ \text{for} \ i \ne j, \nonumber \\
		& \text{and for every} \ i \in \N_p, \ U_i = \{ c_j \}_{ j \in \N_k, j \notin \{ l_i, \ldots, l_p \} }, \nonumber \\ 
		& c_{l_i} \in S \setminus U_i \ \text{is sampled according to} \ P_{U_i} \}; \nonumber
	\end{align}
	
	\item The local search procedure follows Algorithm~\ref{alg:bi_loc_search}. The neighborhood structure $N(C)$ for incumbent solution $C$ is defined as
	$$
	N(C) = \bigcup_{i \in \N_k} \text{Conv}(S_i),
	$$
	where $S_i \subset S, \ i \in \N_k,$ stands for the set of all sample points belonging to cluster $c_i$, i.e., $S_i = \{ x \in S \ | \ \argmin_{c \in C} d(x, c) = c_i \}$, while $\text{Conv}(\cdot)$ is the operation of taking the convex hull of a set. For each cluster, searching for the best improving candidate solution inside the convex hull of all points inside the cluster is justified due to an important property of the squared Euclidean distance function. If a centroid is placed outside the convex hull of the points, the objective function is guaranteed to decrease; whereas, the closer the centroid is placed to the mean of cluster points, the smaller the squared Euclidean distance from the points to the centroid becomes~\cite{Cuong2020}. It is not hard to show that the maximum improvement of the objective function~\eqref{eq:mssc} is attained by placing new centroids at the means of the clusters~\cite{Cuong2020}. K-means is based exactly on this crucial property. Hence, K-means is used as the local search procedure in the proposed algorithm;
	
	\item The neighborhood change step follows Algorithm~\ref{alg:cyc_nbhd_change}. In each iteration, the shaking power $p$ is incremented irrespective of whether the objective function has decreased or not.
\end{enumerate}

It has been established empirically that it is exactly the above choice of VNS ingredients that guarantees the best performance of BigVNSClust both with respect to the resulting time and quality of the obtained solution.

It should be noted that BigVNSClust does not strictly adhere to the VNS metaheuristic, as both the neighborhood structure $\NN$ for shaking and the local search procedure are applied to a sample $S$ rather than the entire data space $X$. Nonetheless, the article~\cite{Mussabayev2024-vls} introduced a new metaheuristic, Variable Landscape Search (VLS), under which BigVNSClust is categorically included.

\subsection{Analysis of the algorithm}

The proposed algorithm iteratively builds and optimizes partial solution landscapes of the MSSC problem restricted to relatively small input data samples $S$ randomly drawn from original dataset $X$. The K-means algorithm plays the role of a local search for identifying best solutions in cyclically expanding neighborhoods of incumbent solution $C$.

Also, BigVNSClust employs an explicit procedure for shaking the incumbent solution. After the local search phase, exactly $p$ random centroids, including any degenerate ones, are reinitialized in the incumbent solution $C$. Clearly, as $p$ increases, the perturbation applied to $C$ becomes more intense. In the edge case when $p_{\max} = k$ and $p$ reaches $p_{\max}$, the whole incumbent solution is reinitialized according to the K-means++ logic. This amounts to a total restart. However, it is most practical to set $p_{\max}$ much smaller than $k$. This limits the strength of applied perturbations, and therefore helps to avoid a complete or near-complete restart.

The shaking power $p$ is increased in every iteration. This allows to incrementally, but in a limited manner, expand the search space for the local search procedure in each new partial solution landscape. By employing the K-means++ logic to sample new centroids for every shaken incumbent solution, we ensure that the new centroids are distributed throughout the sample for optimal coverage. This strategy mitigates the risk of the incumbent solution settling into local minima where multiple cluster centers are positioned so closely that they incorrectly segment a single true cluster.

\subsection{Time complexity}

In Algorithm~\ref{alg:big_vns_clust}, the K-means++ reinitialization of $p$ and all degenerate clusters (as seen in line 9) shares the same time complexity, $\O(s \cdot n \cdot k)$, with a single iteration of the K-means local search. This equivalence arises since all operations are executed on a sample $S$ of size $s$. The additional segment in lines 19 to 22 of Algorithm~\ref{alg:big_vns_clust}, which deals with the incrementation of shaking power $p$, operates at a complexity of $\O(1)$. Consequently, the overall time complexity of a single iteration of the BigVNSClust method remains $\O(s \cdot n \cdot k)$, consistent with the preceding Big-means algorithm.

\section{Experimental evaluation} \label{sec:big_vns_clust_experiments}

\subsection{Hardware and software}

We conducted our experiments on a system running Ubuntu 22.04 64-bit, powered by an AMD EPYC 7663 processor with 8 active cores. The system had 1.46 TB of RAM. Our software setup included Python 3.10.11, NumPy 1.24.3, and Numba 0.57.0. We used Numba~\cite{Marowka2018} to speed up Python code execution and enable parallel processing. Numba's ability to compile Python code into machine code and run it on multiple processors made it a valuable tool for our experiments.

\subsection{Competitive algorithms}

A detailed experimental analysis is performed to quantify the enhancements brought by BigVNSClust in comparison with the state-of-the-art Big-means algorithm~\cite{Mussabayev2023}. To ensure a fair and focused comparison, we evaluate both algorithms under the same conditions, employing the inner parallelization technique, which was proposed in \cite{Mussabayev2024-hybrid}. Investigating the optimal parallelization approach for the BigVNSClust algorithm might be a subject for future research.

\subsection{Datasets}

We conducted experiments on 23 datasets, comprising 19 publicly available and 4 normalized datasets. These datasets are identical to those used in~\cite{Mussabayev2023}, where additional information and URLs can be accessed. Our datasets are numerical, with no missing values, and exhibit significant diversity in terms of size (ranging from 7,797 to 10,500,000 instances) and attribute count (varying from 2 to 5,000). This diversity allows us to evaluate BigVNSClust's adaptability across various data scales. Furthermore, we followed the methodology of Karmitsa et al.~\cite{Karmitsa2018} to facilitate comparative analysis.

\subsection{Experimental design and evaluation metrics}

We performed clustering on each dataset $n_{exec}$ times, using different numbers of clusters ($k$): 2, 3, 5, 10, 15, 20, and 25. Each clustering run was considered a separate experiment. We assessed each experiment by measuring the relative error ($\eps$) and CPU time ($t$). The relative error metric compared the algorithm's performance to the historical best results ($f^*$), using the formula $\eps = 100 \cdot (f - f^*) / f^*$. If the relative error is negative, it means the algorithm performed even better than expected.

Our experimental results are organized in a custom table format. Each algorithm and the combination of a dataset with a cluster number ($X$, $k$) generates a set of $n_{exec}$ experiments. We computed the minimum, median, and maximum relative accuracy $\eps$ and time $t$ values for each set, averaging across multiple runs. The tables show the average values of these metrics for each dataset, aggregated across various $k$ values. Tables~\ref{tab:big_vns_clust_result_e} and \ref{tab:big_vns_clust_result_t} provide a comparison between BigVNSClust and Big-means.

For example, consider a table entry for a specific algorithm: ISOLET \#Succ = 6/7; Min = 0.01; Median = 0.24; Max = 0.59. This entry means that for each of the 7 different cluster numbers (k = 2, 3, 5, 10, 15, 20, and 25), we ran multiple experiments. For each combination of dataset and cluster number, we performed 15 independent runs of each algorithm. In total, we have 7 sets of runs for each algorithm, with each set containing 15 results. The ``\#Succ'' value of 6/7 indicates that, for 6 out of 7 sets, the median performance of this algorithm was better than the median performance of all other algorithms.

The last rows of the tables summarize the overall performance of each algorithm across all datasets. We bolded the top-performing results for each metric and dataset pair to highlight the best algorithms. An algorithm is deemed successful when its median performance for a given cluster number ($k$) matches or surpasses the best result among all algorithms for that cluster number.

\subsection{Hyperparameter selection}

We set a CPU time limit ($T$) and stopped the K-means clustering process if it took too long (over 300 iterations) or made very little progress (less than 0.0001 improvement between steps). For K-means++, we chose three candidate points for each new centroid. We fine-tuned sample sizes based on initial tests to ensure optimal performance. The exact values for $T$ and $n_{exec}$ are listed in the supplementary material's detailed tables.

\subsection{Experimental results}

The total number of conducted experiments reached 7,366.

In our experimentation, we did not detect significant correlation between the obtained results and the choice of a neighborhood change procedure. However, the cyclic neighborhood change procedure produced slightly better final accuracy $\eps$ than the sequential one.

Also, we explored two distinct methodologies for centroid shaking within our algorithm: a uniformly random approach and a reinitialization based on the K-means++ strategy. It was observed that, while the uniformly random approach can be executed more rapidly than its K-means++ counterpart, it resulted in a marked decrease in final accuracy, measured by $\eps$. Specifically, the accuracy deteriorated by up to fivefold in comparison to the K-means++ reinitialization. This significant discrepancy in performance can be attributed to the nature of the uniformly random method, which introduces an excessive level of perturbation to the existing solution. Such a drastic alteration necessitates a considerably higher number of iterations in the subsequent local search phase to converge to an optimal solution.

For every algorithm, dataset $X$, and number of clusters $k$, the minimum, median, and maximum values of relative accuracy $\eps$ and CPU time $\overline{t}$ were calculated with respect to $n_{exec}$ runs. To determine the optimal value of the maximum shaking power $p_{\max}$, all the experiments were restarted with different values of $p_{\max}$. The resulting performance of the proposed BigVNSClust algorithm across various values of $p_{\max}$ is shown in Table~\ref{tab:big_vns_clust_result_e_t_var_p_max}.

\begin{table}[!htbp]%
	\centering%
	%	\captionsetup{font=scriptsize}
	\caption{Summarized performance of the proposed BigVNSClust algorithm with varying values of the maximum shaking power $p_{\max}$}%
	\label{tab:big_vns_clust_result_e_t_var_p_max}%
	\begin{tabular}{l|cccc|cccc}
		\hline
		\multirow{2}{*}{$p_{\max}$} & \multicolumn{4}{p{3cm}}{\mbox{Relative accuracy $\eps$}} & \multicolumn{4}{p{3cm}}{\mbox{Time $t$}} \\
		\cline{2-9}
		& \#Succ & Min & Median & Max & \#Succ & Min & Median & Max  \\
		\hline
		2 & 114/165 & 0.12 & 0.78 & \textbf{4.51} & 60/165 & \textbf{0.86} & 3.12 & \textbf{4.77} \\
		3 & 112/165 & \textbf{0.08} & 0.72 & 1784.89 & 59/165 & 0.93 & 3.06 & 4.94 \\
		4 & \textbf{122/165} & 0.09 & \textbf{0.67} & 3000.59 & 50/165 & 0.95 & \textbf{3.0} & 4.84 \\
		5 & 108/165 & 0.1 & 0.68 & 5.67 & 50/165 & 0.89 & 3.05 & 5.02 \\
		\hline
	\end{tabular}
\end{table}

Table~\ref{tab:big_vns_clust_result_e_t_var_p_max} indicates that more intensive shaking leads to improved accuracy and a slightly increased maximal convergence time. The optimal value of $p_{\max}$, in terms of median relative accuracy, is $4$. From these results, it can be inferred that more vigorous shaking enhances the incumbent solution's ability to escape local minima by ``jumping out'' of their valleys. However, as anticipated, descending from these perturbed positions using local search incurs additional time.

For the best-performing configuration of BigVNSClust with parameter value $p_{\max} = 4$, the overall results are shown in Tables~\ref{tab:big_vns_clust_result_e} -- \ref{tab:big_vns_clust_result_t}.

\begin{table}[!htbp]%
	\centering%
	%	\captionsetup{font=scriptsize}
	\caption{Relative clustering accuracies $\epsilon$ (in \%) resulting from the comparison of BigVNSClust with Big-means}%
	\label{tab:big_vns_clust_result_e}%
%	\resizebox{0.8\linewidth}{!}{%
	\resizebox{!}{1.3\summtableheight}{%
		\begin{tabular}{l|cccc|cccc}
			\hline
			\multirow{2}{*}{Dataset} & \multicolumn{4}{p{3cm}}{\mbox{BigVNSClust}}& \multicolumn{4}{p{3cm}}{\mbox{Big-means}} \\
			\cline{2-9}
			& \#Succ & Min & Median & Max & \#Succ & Min & Median & Max  \\
			\hline
			CORD-19 Embeddings & 4/7 & \textbf{0.04} & \textbf{0.12} & \textbf{0.26} & 3/7 & 0.06 & 0.27 & 0.44 \\
			HEPMASS & 5/7 & \textbf{0.07} & \textbf{0.14} & \textbf{0.26} & 2/7 & 0.09 & 0.27 & 1.05 \\
			US Census Data 1990 & 7/7 & \textbf{0.4} & \textbf{1.39} & \textbf{2.76} & 0/7 & 0.89 & 3.17 & 33.23 \\
			Gisette & 2/7 & \textbf{-0.44} & -0.36 & \textbf{-0.25} & 5/7 & \textbf{-0.44} & \textbf{-0.38} & -0.15 \\
			Music Analysis & 4/7 & 0.34 & \textbf{0.91} & \textbf{2.23} & 3/7 & \textbf{0.32} & 1.16 & 7.48 \\
			Protein Homology & 4/7 & 0.38 & \textbf{0.83} & \textbf{1.97} & 3/7 & \textbf{0.14} & 0.87 & 5.31 \\
			MiniBooNE Particle Identification & 5/7 & \textbf{-0.08} & \textbf{0.06} & \textbf{3.07} & 2/7 & -0.02 & 0.63 & 22.53 \\
			MiniBooNE Particle Identification (normalized) & 3/7 & 0.28 & 0.68 & \textbf{1.57} & 4/7 & \textbf{0.16} & \textbf{0.54} & 3.23 \\
			MFCCs for Speech Emotion Recognition & 5/7 & \textbf{0.15} & \textbf{0.42} & \textbf{0.96} & 2/7 & 0.19 & 0.87 & 1.43 \\
			ISOLET & 5/7 & \textbf{-0.02} & \textbf{0.29} & \textbf{0.89} & 2/7 & 0.19 & 0.61 & 1.66 \\
			Sensorless Drive Diagnosis & 7/7 & -0.4 & \textbf{0.15} & \textbf{13.07} & 0/7 & \textbf{-0.43} & 3.68 & 40.41 \\
			Sensorless Drive Diagnosis (normalized) & 6/7 & \textbf{0.45} & \textbf{1.8} & \textbf{4.92} & 1/7 & 0.64 & 2.39 & 7.9 \\
			Online News Popularity & 6/7 & \textbf{0.57} & \textbf{1.78} & \textbf{6.42} & 1/7 & 0.95 & 2.97 & 22.41 \\
			Gas Sensor Array Drift & 7/7 & \textbf{-0.04} & \textbf{0.74} & \textbf{3.83} & 0/7 & 0.15 & 3.23 & 9.6 \\
			3D Road Network & 2/7 & \textbf{0.04} & \textbf{0.39} & 1.7 & 5/7 & 0.05 & 0.42 & \textbf{1.21} \\
			Skin Segmentation & 6/7 & \textbf{0.21} & \textbf{1.93} & \textbf{7.02} & 1/7 & 0.25 & 3.56 & 10.32 \\
			KEGG Metabolic Relation Network (Directed) & 5/7 & \textbf{-0.38} & \textbf{0.35} & \textbf{3.13} & 2/7 & -0.37 & 2.2 & 32.84 \\
			Shuttle Control & 8/8 & \textbf{-0.75} & \textbf{0.08} & \textbf{6.07} & 0/8 & 0.11 & 5.59 & 32.58 \\
			Shuttle Control (normalized) & 6/8 & \textbf{1.15} & 3.15 & \textbf{8.1} & 2/8 & 1.16 & \textbf{2.78} & 16.0 \\
			EEG Eye State & 8/8 & \textbf{-0.02} & \textbf{0.01} & 68938.59 & 0/8 & 0.54 & 2.64 & \textbf{7.02} \\
			EEG Eye State (normalized) & 8/8 & \textbf{-0.06} & \textbf{-0.03} & \textbf{5.34} & 0/8 & \textbf{-0.06} & 8.74 & 36.58 \\
			Pla85900 & 4/7 & 0.1 & \textbf{0.31} & \textbf{0.93} & 3/7 & \textbf{0.08} & 0.43 & 1.44 \\
			D15112 & 5/7 & \textbf{0.1} & \textbf{0.25} & \textbf{0.75} & 2/7 & 0.11 & 0.52 & 4.33 \\
			\hline
			Overall Results & 122/165 & \textbf{0.09} & \textbf{0.67} & 3000.59 & 43/165 & 0.21 & 2.05 & \textbf{12.99} \\ \hline
		\end{tabular}%
	}
\end{table}

\begin{table}[!htbp]%
	\centering%
	%	\captionsetup{font=scriptsize}
	\caption{Processing times $t$ (in seconds) resulting from the comparison of BigVNSClust with Big-means}%
	\label{tab:big_vns_clust_result_t}%
	%	\resizebox{0.8\linewidth}{!}{%
	\resizebox{!}{1.3\summtableheight}{%
		\begin{tabular}{l|cccc|cccc}
			\hline
			\multirow{2}{*}{Dataset} & \multicolumn{4}{p{3cm}}{\mbox{BigVNSClust}}& \multicolumn{4}{p{3cm}}{\mbox{Big-means}} \\
			\cline{2-9}
			& \#Succ & Min & Median & Max & \#Succ & Min & Median & Max  \\
			\hline
			CORD-19 Embeddings & 4/7 & \textbf{8.94} & \textbf{22.78} & 35.38 & 3/7 & 10.05 & 23.75 & \textbf{35.36} \\
			HEPMASS & 2/7 & 6.15 & 19.55 & 28.88 & 5/7 & \textbf{3.28} & \textbf{16.97} & \textbf{27.86} \\
			US Census Data 1990 & 1/7 & 0.28 & 1.99 & 3.09 & 6/7 & \textbf{0.24} & \textbf{1.59} & \textbf{2.96} \\
			Gisette & 5/7 & \textbf{3.03} & \textbf{4.09} & 7.53 & 2/7 & 3.05 & 4.82 & \textbf{7.29} \\
			Music Analysis & 5/7 & \textbf{0.27} & \textbf{3.85} & 8.09 & 2/7 & 0.63 & 4.59 & \textbf{7.88} \\
			Protein Homology & 6/7 & \textbf{0.32} & \textbf{1.84} & 3.44 & 1/7 & 0.57 & 2.09 & \textbf{3.39} \\
			MiniBooNE Particle Identification & 3/7 & \textbf{0.69} & 2.2 & 3.94 & 4/7 & 0.82 & \textbf{1.96} & \textbf{2.99} \\
			MiniBooNE Particle Identification (normalized) & 1/7 & 0.14 & 0.57 & 1.06 & 6/7 & \textbf{0.1} & \textbf{0.47} & \textbf{0.98} \\
			MFCCs for Speech Emotion Recognition & 4/7 & \textbf{0.08} & 0.57 & 1.08 & 3/7 & 0.14 & \textbf{0.55} & \textbf{0.99} \\
			ISOLET & 3/7 & \textbf{0.39} & 3.02 & 4.8 & 4/7 & 0.71 & \textbf{2.74} & \textbf{4.79} \\
			Sensorless Drive Diagnosis & 0/7 & 0.19 & 0.81 & 1.36 & 7/7 & \textbf{0.17} & \textbf{0.65} & \textbf{1.03} \\
			Sensorless Drive Diagnosis (normalized) & 0/7 & 0.05 & 0.21 & 0.34 & 7/7 & \textbf{0.02} & \textbf{0.16} & \textbf{0.29} \\
			Online News Popularity & 2/7 & 0.09 & 0.43 & 0.73 & 5/7 & \textbf{0.07} & \textbf{0.36} & \textbf{0.68} \\
			Gas Sensor Array Drift & 0/7 & 0.13 & 1.26 & 2.07 & 7/7 & \textbf{0.12} & \textbf{0.98} & \textbf{1.97} \\
			3D Road Network & 0/7 & \textbf{0.08} & 0.38 & 0.69 & 7/7 & \textbf{0.08} & \textbf{0.31} & \textbf{0.54} \\
			Skin Segmentation & 0/7 & 0.05 & 0.17 & 0.25 & 7/7 & \textbf{0.03} & \textbf{0.11} & \textbf{0.2} \\
			KEGG Metabolic Relation Network (Directed) & 0/7 & 0.16 & 0.72 & 1.15 & 7/7 & \textbf{0.08} & \textbf{0.54} & \textbf{0.99} \\
			Shuttle Control & 2/8 & 0.28 & 1.04 & 1.54 & 6/8 & \textbf{0.07} & \textbf{0.77} & \textbf{1.44} \\
			Shuttle Control (normalized) & 1/8 & 0.06 & 0.25 & 0.41 & 7/8 & \textbf{0.03} & \textbf{0.21} & \textbf{0.38} \\
			EEG Eye State & 2/8 & 0.16 & 0.96 & 1.52 & 6/8 & \textbf{0.07} & \textbf{0.8} & \textbf{1.47} \\
			EEG Eye State (normalized) & 3/8 & 0.09 & 0.68 & 1.05 & 5/8 & \textbf{0.04} & \textbf{0.56} & \textbf{0.99} \\
			Pla85900 & 1/7 & 0.11 & 0.84 & 1.52 & 6/7 & \textbf{0.04} & \textbf{0.76} & \textbf{1.46} \\
			D15112 & 5/7 & 0.11 & \textbf{0.74} & 1.42 & 2/7 & \textbf{0.09} & 0.78 & \textbf{1.41} \\
			\hline
			Overall Results & 50/165 & 0.95 & 3.0 & 4.84 & 115/165 & \textbf{0.89} & \textbf{2.89} & \textbf{4.67} \\ \hline
		\end{tabular}%
	}
\end{table}

\subsection{Synthetic experiment} \label{subsec:big_vns_clust_synth_expers}

In every BigVNSClust iteration, the sampling process sparsifies the given dataset to some extent. We anticipate that this sparsification enables the initial centroids to shift more freely between the clusters, provided the clusters overlap. In essence, sampling with a carefully selected sample size enhances the flexibility of centroids within the current solution while still maintaining a reasonable approximation of the original dataset. Consequently, it is crucial to strike the appropriate balance between these two objectives when determining the sample size.

To examine this hypothesis further, we have conducted additional experiments wherein we assess the performance of BigVNSClust using carefully generated synthetic data.

\begin{figure}[htb]
	\captionsetup[subfigure]{justification=centering}
	\centering
	\begin{subfigure}[b]{0.32\textwidth}
		\centering
		\includegraphics[scale=0.2]{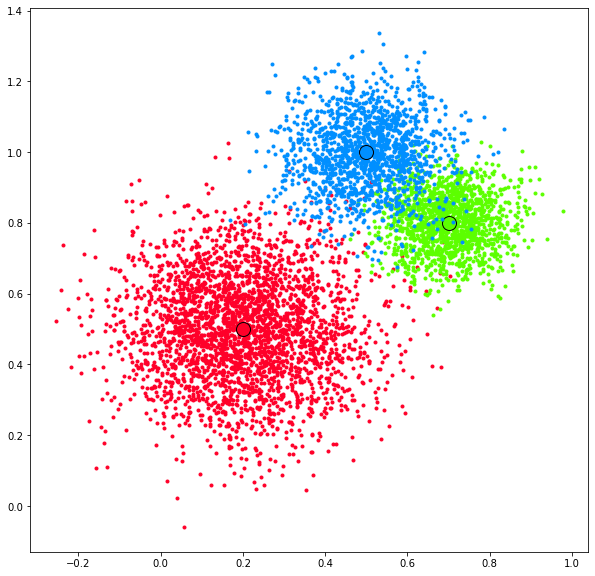}
		\caption{}
		\label{fig_x1}
	\end{subfigure}%
	\begin{subfigure}[b]{0.32\textwidth}
		\centering
		\includegraphics[scale=0.2]{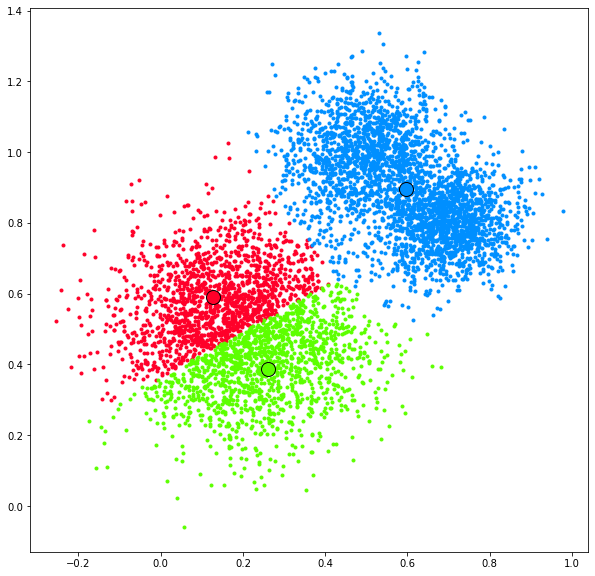}
		\caption{}
		\label{fig_kmpp_x1}
	\end{subfigure}%
	\begin{subfigure}[b]{0.32\textwidth}
		\centering
		\includegraphics[scale=0.2]{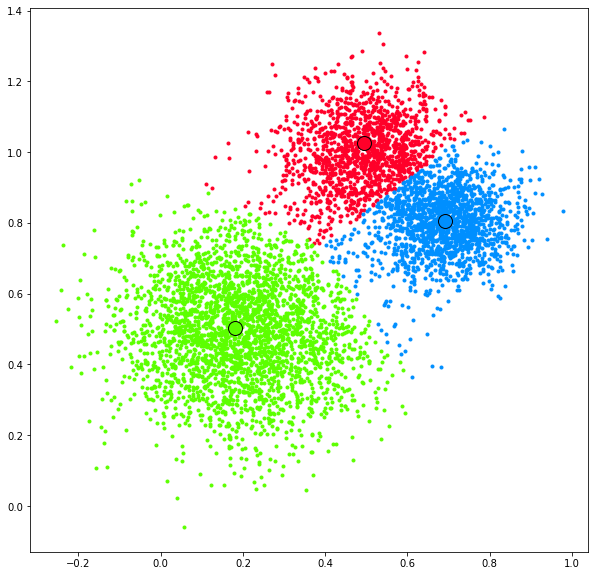}
		\caption{}
		\label{fig_bm_x1}
	\end{subfigure}
	\caption{In \ref{fig_x1}: the original dataset $X_1$; in \ref{fig_kmpp_x1}: the result of running K-means++ on $X_1$ with $C$ as initial centroids; in \ref{fig_bm_x1}: the result of running BigVNSClust on $X_1$ with $C$ as initial centroids}
\end{figure}

First, consider a mixture of 3 isotropic Gaussian distributions $\{ \NN_1, \NN_2, \NN_3 \}$ with standard deviations $0.15, 0.08, 0.1$, respectively. Let $\mu_1 = [0.2, 0.5]^T, \mu_2 = [0.7, 0.8]^T, \mu_3 = [0.5, 1.0]^T$ be the coordinates of the means of these distributions. Let $X_1$ be the dataset obtained by sampling 3000 points from $\NN_1$, 1500 points from $\NN_2$, and 1500 points from $\NN_3$. Dataset $X_1$ is shown in Figure \ref{fig_x1}. If we assign the initial centroids $C = \{ c_1, c_2, c_3 \}$ as $c_1 = [0.1, 0.2]^T, c_2 = [0.1, 0.15]^T, c_3 = [0.5, 1.0]^T$, then the result of running K-means++ on $X_1$ with $C$ as initial centroids is shown in Figure \ref{fig_kmpp_x1}. The value of the objective function $f(\{ \mu_1, \mu_2, \mu_3 \}, X_1)$ for the ground truth is $171.5$. We see that K-means++ was trapped in a local minimum with the objective value of $191.52$. Choosing the sample size $s = 70$ and $T = 1.5$, it is possible to obtain a result of BigVNSClust as shown in Figure \ref{fig_bm_x1}, with the objective value equal to $172.08$. Hence, BigVNSClust managed to avoid being trapped in a bad local minimum.

\subsection{Analysis of obtained results}

The summarized experimental results reveal that incorporating the VNS metaheuristic into the Big-means scheme leads to a significant boost of the clustering accuracy $\eps$. Concretely, the accuracy increased by more than threefold in comparison to the original Big-means algorithm.

We note that the integration of a new advanced parallelization method could enhance BigVNSClust's performance even further. Exploring such a parallelization method may be an exiting direction for future research.

In spite of the fact that BigVNSClust exhibited slightly worse results than Big-means with respect to the processing time $t$, its time results are still much better than those of all other advanced parallel HPClust versions from \cite{Mussabayev2024-hybrid}. One must also acknowledge the fact that the current average time result of $3.0$ seconds achieved by BigVNSClust is practically acceptable in most cases. Thus, BigVNSClust can be readily recommended for immediate use by practitioners as a powerful tool for solving industry-level big data clustering problems.

The higher resulting accuracy of BigVNSClust is attributed to the novel shaking properties introduced by the VNS metaheuristic. This iterative shaking of the incumbent solution addresses a key issue with Big-means, discussed in Subsection~\ref{subsec:big_vns_clust_synth_expers}. Specifically, BigVNSClust inherently possesses a natural shaking property due to its iterative sampling approach. The addition of VNS further enhances this by iteratively reassigning a portion of the incumbent centroids. This process not only shakes the solution landscape more intensely but also allows centroids from distinct, non-overlapping clusters to transition between each other's clusters, facilitating better solution exploration.

It is worth noting that the advanced parallel schemes \cite{Mussabayev2024-hybrid} also naturally tackle the aforementioned problem, thus also achieving better accuracy results that Big-means. Specifically, each advanced parallel scheme employs a unique tactic to increase the likelihood of selecting a centroid configuration where no multiple centroids are placed within a single well-separated cluster. For example, HPClust-competitive accomplishes this by running numerous independent clustering processes with various initializations and ultimately selecting the best result. HPClust-cooperative does so by thoroughly exploring the optimal initial configuration from a vast array. Lastly, HPClust-hybrid effectively combines these two approaches.

\section{Conclusion and future research} \label{sec:big_vns_clust_conclusion}

In this work, we proposed BigVNSClust, a novel VNS-based and big-data-efficient MSSC algorithm. Also, the basics of VNS were reviewed, while the most efficient choice of VNS ingredients has been empirically established to ensure superior algorithm performance. The conducted experimental analysis confirmed that BigVNSClust provides a significant increase in the obtained accuracy over the state-of-the-art Big-means approach, being on par in this performance metric with other advanced HPClust parallel schemes (competitive, cooperative, and hybrid). Also, BigVNSClust showed a much better time efficiency compared to the advanced parallel HPClust versions, while being only slightly inferior to HPClust-inner in that regard.

These results convince us that the VNS metaheuristic favorably combines with the natural shaking properties of the problem decomposition approach, producing a novel algorithm with highly desirable practical qualities. This result was achieved at the cost of minimal algorithmic complexity. Exploring more advanced ways to parallelize BigVNSClust is a promising future research direction that can make the algorithm even more effective and efficient.

More possible future research directions include: the development of a tool that would allow to automatically choose the appropriate parameter triple ($p_{\max}, \ s, \ T$) for each individual dataset, extension of the algorithm to the clustering paradigms other than MSSC, and exploring combinations of other well-known metaheuristics that can lead to an even more significant improvement of Big-means.

\section*{Acknowledgements}

This research was funded by the Science Committee of the Ministry of Science and Higher Education of the Republic of Kazakhstan (grant no. BR21882268).

\bibliography{main}

\end{document}